\documentclass[runningheads]{llncs}
\usepackage[T1]{fontenc}
\usepackage{graphicx}
\usepackage{booktabs}
\usepackage[misc]{ifsym}

\usepackage{mwe}
\usepackage{caption} 
\usepackage{amssymb}
\usepackage{booktabs}
\usepackage{xcolor}
\usepackage{pifont}
\usepackage{algorithm}
\usepackage{algorithmic}
\usepackage{newfloat}
\usepackage{listings}
\usepackage{multirow}
\usepackage{tabularx}
\usepackage{amsmath}
\usepackage{graphicx}
\usepackage{subfig}
\usepackage{soul}
\usepackage{graphicx}
\usepackage{pifont}
\usepackage{enumitem}

\begin{document}

\title{DiffuSent: Towards a Unified Diffusion Framework for Aspect-Based Sentiment Analysis}

\titlerunning{DiffuSent: Diffusion Framework for Aspect-Based Sentiment Analysis}


\author{Shu Long$^\star$\inst{1} \and Yanglei Gan\thanks{Equal contribution.}\inst{2} \inst{3} \and Xuchuan Zhou \inst{2}~(\Letter)}

\authorrunning{Long et al.}

\institute{Southwest Petroleum University \email{202199010018@swpu.edu.cn}\and Southwest Minzu University \email{xczhou@swun.edu.cn} \email{yangleigan@swun.edu.cn} \and University of Electronic Science and Technology of China}


\toctitle{DiffuSent: Towards a Unified Diffusion Framework for Aspect-Based Sentiment Analysis}
\tocauthor{Shu Long, Yanglei Gan, Xuchuan Zhou}

\maketitle              

\begin{abstract}
Aspect-Based Sentiment Analysis (ABSA) encompasses seven distinct subtasks, each focusing on different extracted elements. Despite the proven success of generative models in unified aspect sentiment analysis, existing approaches often rely on auto-regressive token-by-token generation without grasping the whole information of the aspect and opinion terms, resulting in boundary insensitivity, particularly in context of multi-word aspect and opinion terms. To address these issues, we present DiffuSent, a non-auto-regressive diffusion framework that systematically formulates all ABSA subtasks as boundary denoising diffusion processes, progressively refining boundaries over noisy states. Furthermore, we introduce a contrastive denoising training strategy which effectively address duplicate predictions with subtle variations introduced by diffusion process. Extensive experiments across 28 settings (7 subtasks $\times$ 4 datasets) demonstrate that DiffuSent achieves delivers consistent improvements over the strongest generative and span-based systems. DiffuSent exhibits notable gains on multi-word triplets, achieving an average improvement of +2.48 F1, and maintains robust extraction accuracy in sentences containing multiple sentiment triplets. Moreover, the non-auto-regressive decoding enables substantial efficiency benefits, reaching up to 181$\times$ faster inference than auto-regressive generative baselines.

\end{abstract}

\section{Introduction}
\begin{figure}[t]
    \centering
\includegraphics[width=0.7\textwidth]{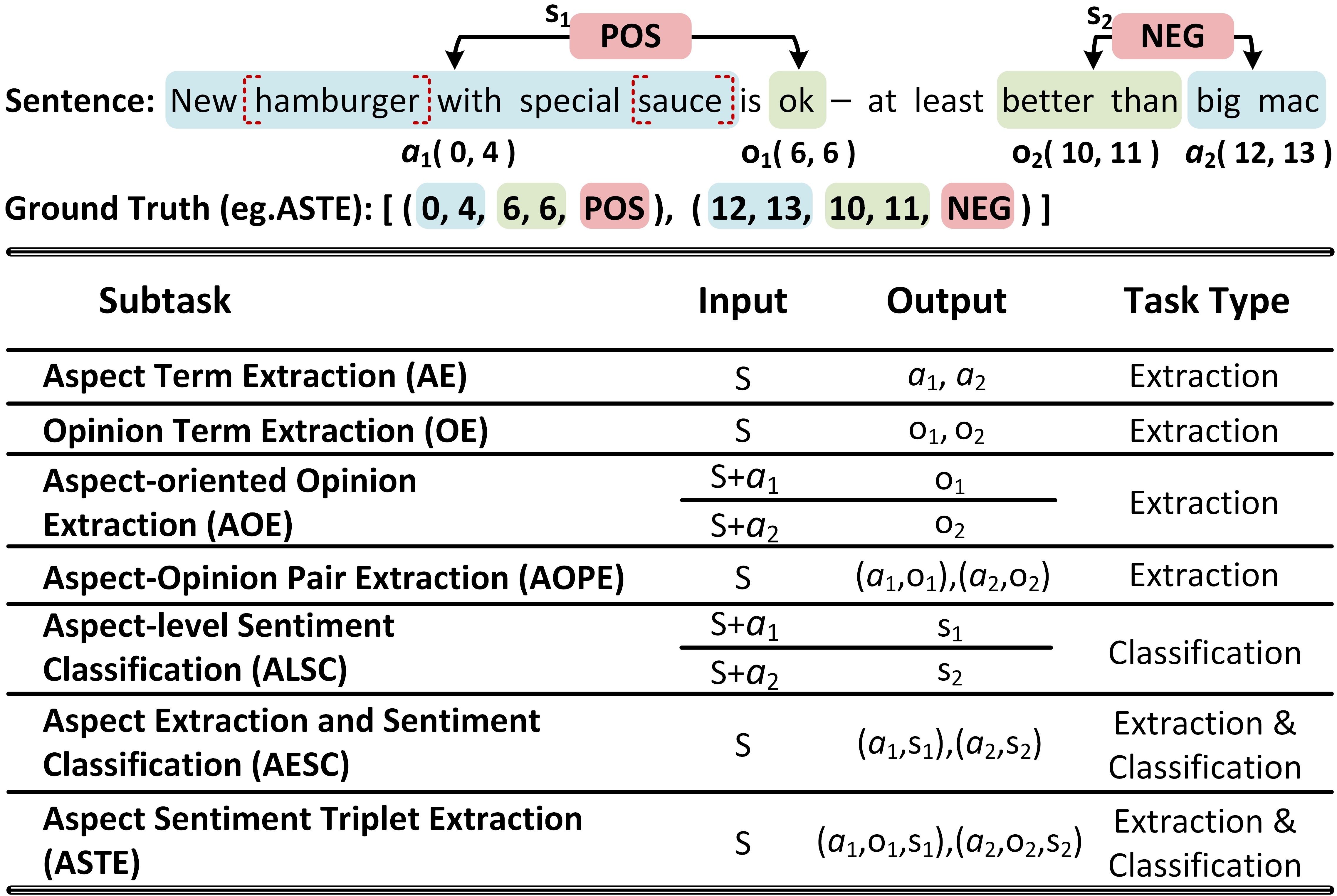}
    \caption{Illustration of seven ABSA subtasks}
    \label{fig:task_definition}
\end{figure}

Aspect-based Sentiment Analysis (ABSA) stands as a fine-grained branch of sentiment analysis that shifts the focus from coarse document- or sentence-level sentiment to sentiment expressed toward specific entities \cite{pontiki2016semeval}. An ABSA instance is typically composed of three essential elements: an aspect term (\textit{a}), an opinion term (\textit{o}), and a sentiment polarity (\textit{s}). To illustrate, consider the review sentence from {\tt Res 15} shown in Figure \ref{fig:task_definition}, "\textit{New hamburger with special sauce is ok – at least better than big mac}". The phrases "\textit{New hamburger with special sauce}" and "\textit{big mac}" function as aspect terms, while "\textit{ok}" and "\textit{better than}" act as opinions linked to positive/negative sentiment polarities, respectively. 

Conventional approaches to ABSA have focused on distinct components such as aspect/opinion term extraction \cite{ma2019exploring,zhao2020spanmlt}, sentiment classification for a given aspect \cite{tang2015document,liu2023Aspect}, or aspect sentiment triplet extraction \cite{mukherjee2021paste,zhang2022boundary,zhou-qian-2023-strength,lin2025revisiting}. While these developments have led to successes in individual subtasks, a unified ABSA framework remains an elusive goal. To bridge this gap, recent research has been shifting towards unified approaches within a pipeline framework \cite{mao2021joint,fei2022inheriting}. However, such paradigms often suffer from error accumulation due to their modular approaches \cite{Fei21Nonautoregressive}. Addressing these drawbacks, there is a growing inclination towards employing generative models in ABSA. This shift signifies a move to an end-to-end autoregressive formulation, broadening the scope to include techniques such as word index generation \cite{yan-etal-2021-unified}, label augmented text generation \cite{zhang-etal-2021-towards-generative}, and template filling \cite{gao2022lego,gou-etal-2023-mvp}. Despite their effectiveness, auto-regressive generative models suffer from two critical drawbacks:

\begin{itemize}[leftmargin=*]
    \item \textbf{Boundary Ambiguity and Integrity Failure in Multi-Word Extraction.} The step-wise, token-focused mechanism of auto-regressive decoding fundamentally compromises the integrity of multi-word aspect and opinion terms. Since the model generates output sequentially, conditioning each token primarily on immediate history \cite{cao2021image,zhou-qian-2023-strength,gan2025optimizing}, it restricts the capacity to capture the holistic context necessary for complete multi-word expressions. This often results in boundary ambiguity, where the decoder prematurely truncates a descriptive phrase, such as truncating "\textit{new hamburger with special sauce}" to just "\textit{hamburger}". This limitation is especially acute for long, syntactically complex, or highly descriptive aspect terms.
    \item \textbf{Computational Bottleneck Due to Sequential Decoding.} The inherent requirement for strictly sequential token generation in auto-regressive models imposes a significant computational inefficiency. The decoding process enforces strict temporal dependencies, which fundamentally precludes parallel computation across the sequence length. Consequently, the inference latency scales linearly with the length of the target structured output \cite{Fei21Nonautoregressive,xiao2023survey}. This computational cost becomes prohibitive when generating lengthy, structured outputs like multiple aspect–opinion pairs. Ultimately, this bottleneck limits the scalability of generative ABSA systems in real-world applications.
\end{itemize}

To bridge this gap, we propose DiffuSent, a novel unified generative diffusion framework tailored for ABSA. Distinct from traditional token-by-token generation paradigm, DiffuSent is designed to explicitly model boundary indices, and dynamically refines its interpretations based on comprehensive contextual information. Through a non-auto-regressive boundary denoising diffusion process, it delivers predictions for boundary indices in a single step. Specifically, we systematically infuse uncertainty via Gaussian noise into the aspect/opinion term boundaries using a forward diffusion process. The subsequent reverse diffusion process then meticulously refines these term boundaries from their initially indeterminate states. Additionally, we introduce a contrastive denoising training strategy designed to differentiate between accurate and inaccurate boundary predictions. It adeptly manages the duplicate predictions with subtle variations in boundary detection, particularly in distinguishing semantically similar terms such as "\textit{hamburger}", "\textit{new hamburger}", and "\textit{new hamburger with special sauce}". We validate DiffuSent on four benchmarks for seven subtasks and DiffuSent yields state-of-the-art performance. Our contributions are three-fold:

\begin{itemize}[leftmargin=*]
    \item We propose DiffuSent, a unified diffusion-based framework that formulate all ABSA subtasks as boundary denoising diffusion process.
    \item A novel contrastive denoising training strategy is introduced. This strategy is designed to address duplicate predictions with subtle variations in predicted boundary indices introduced by diffusion process.
    \item Extensive experiments are conducted on 28 subtasks (7 $\times$ 4 datasets) to evaluate the effectiveness of our approach. Experimental results demonstrate that our model outperforms the state-of-the-art methods.
\end{itemize}

\section{Related Work}

\paragraph{Aspect-Based Sentiment Analysis.}
Aspect-Based Sentiment Analysis (ABSA) encompasses a suite of interrelated subtasks, each focusing on specific components or their combinations within a text as illustrated in Figure \ref{fig:task_definition}. Previous studies mainly focus on individual subtasks \cite{tang2015effective,wang2017coupled,liu2024let}, including Aspect Term Extraction (AE), Opinion Term Extraction (OE), Aspect-level Sentiment Classification (ALSC). Subsequent research shifted towards integrated models that simultaneously extract aspects, opinions, and their corresponding sentiments \cite{fan-etal-2019-target,gao2021question,hu-etal-2019-open}, such as Aspect-oriented Opinion Extraction (AOE), Aspect-Opinion Pair Extraction (AOPE) and Aspect Extraction and Sentiment Classification (AESC). Marking a significant shift in the field, Peng \textit{et al.} \cite{peng2020knowing} introduced the Aspect Sentiment Triplet Extraction (ASTE) task, pioneering a unified approach for extracting aspect, opinion, and sentiment triplets. This approach led to the development of advanced techniques in ABSA, such as table filling \cite{jing2021seeking,zhang2022boundary}, sequence tagging \cite{xu-etal-2020-position,zhou-qian-2023-strength}, and span-based methods \cite{chen2022span,liang2023stage}. However, these methods only focus on individual tasks.

Recent trends in Aspect-Based Sentiment Analysis (ABSA) have seen the emergence of unified methods, such as two-step MRC approach \cite{mao2021joint}. However, this method suffers from error accumulation due to isolated processing. In response, a shift towards end-to-end generative methods has occurred, addressing all ABSA subtasks more effectively. These include approaches like word index generation \cite{yan-etal-2021-unified}, label augmented text generation \cite{zhang-etal-2021-towards-generative}, and template filling \cite{gao2022lego,gou-etal-2023-mvp,zhou-qian-2023-strength}. However, a notable limitation of these generative models is their reliance on auto-regressive, token-by-token decoding. This approach, while effective, does not fully capitalize on the semantics available in multi-word terms and can be inefficient time-wise. In our work, we utilize a diffusion model to facilitate progressive refinements of term boundaries and output all predictions simultaneously in non-auto-regressive manner, effectively addressing complex linguistic structures.

\paragraph{Diffusion Models.}
Diffusion models \cite{pmlr-v37-sohl-dickstein15}, primarily used for continuous data like images and audio \cite{kong2020diffwave,chen2023diffusiondet}, face challenges when applied to the discrete nature of text in NLP. Innovations by Hoogeboom \textit{et al.} \cite{hoogeboom2021argmax} and Austin \textit{et al.} \cite{austin2021structured} have adapted these models for character-level text generation, while researchers further developed methods to bridge the gap between continuous and discrete domains. Among the initial attempts, Li \textit{et al.} \cite{li2022diffusion} first adapts continuous diffusion models for text by adding an embedding step and a rounding step, and designing a training objective to learn the embedding. Gong \textit{et al.} \cite{gong2023diffuseq} proposes a diffusion model designed for sequence-to-sequence (seq2seq) text generation tasks by adding partial noise during the forward process and conditional denoising during the reverse process. In contrast to them, which are based on continuous diffusion models, some works \cite{he2023diffusionbert,shen-etal-2023-diffusionner} instead explore discrete diffusion models to integrate PLMs as the backbone. He \textit{et al.} \cite{he2023diffusionbert} proposes a non-auto-regressive model for text generation by combining pre-trained denoising language models with absorbing-state discrete diffusion models. Notably, 
Shen \textit{et al.} \cite{shen-etal-2023-diffusionner} frame Named Entity Recognition as a boundary denoising process, offering insights into the application of diffusion models in text extraction. Cai \textit{et al.} and Gan \textit{et al.} \cite{cai2024predicting,gan2026negative} cast future link prediction as a conditional denoising process, where noisy target representations are gradually refined under historical temporal contexts. Building on this innovation, we have developed DiffuSent, a unified generative diffusion framework designed to address all ABSA subtasks.

\section{Methodology}
\subsection{Problem Definition} \label{sec 3.1}
Given a sentence $S=\{w_1,w_2,...,w_M\}$, the objective of ASTE is to extract the boundary indices of all conceivable aspect terms, associated opinion expression terms, and their corresponding sentiment polarity labels, denoted as $T={\{(a_i^s, a_i^e, o_i^s, o_i^e, s_i)\}}_{i=1}^N$. The superscripts $s$ and $e$ denote the start and end indices of aspect or opinion terms within the input text. The sentiment polarity label $s_i$ takes values from $\{\tt{POS, NEU, NEG}\}$, and $N$ signifies the count of target triples. We define boundary sequences as $T_b={\{(a_i^s, a_i^e, o_i^s, o_i^e)\}}_{i=1}^N$ to facilitate the subsequent presentation.

\subsection{Boundary Denoising Diffusion Process}
\begin{figure*}[t]
    \centering
    \includegraphics[width=0.95\textwidth]{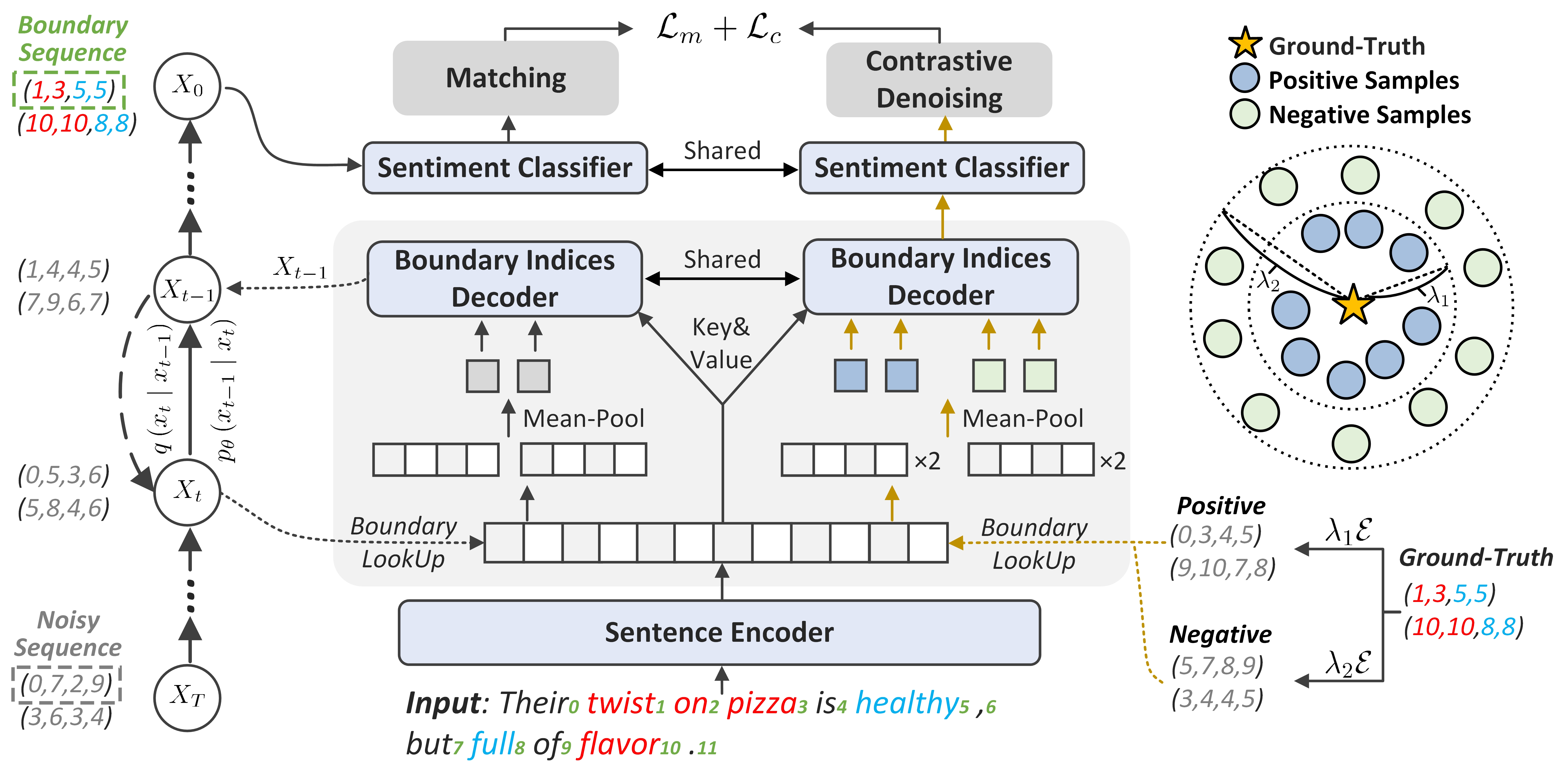}
    \caption{Overview of DiffuSent. "\textit{Boundary LookUp}" denotes get corresponding word embedding with boundary as index. The stream identified with "\textcolor[RGB]{191,144,0}{$\uparrow$}" only occurs in the last reverse process in training stage and does not exist in inference stage. Noise $\mathcal{E} \sim \mathcal{N}(\mathbf{0}, \mathbf{I})$.}
    \label{fig:model overview}
\end{figure*}

As shown in Figure \ref{fig:model overview}, in our boundary denoising diffusion process, the boundary sequences $T_b$ are considered as data samples. During the forward diffusion phase, Gaussian noise is incrementally added to indices in these sequences. Conversely, the reverse diffusion process aims to restore the original boundary indices.

\textbf{Boundary Indices Forward Diffusion.}
In this phase, we progressively introduce Gaussian noise to the boundary sequences $T_b \in \mathbb{R}^{N \times 4}$, simulating the uncertainty inherent in identifying term boundaries. To facilitate parallel training, we normalize the count $N$ of $T_b$ to $N_{train}$ by duplicating, with normalized sequences represented as $\mathbf{x}_0 \in \mathbb{R}^{N_{train} \times 4}$. The noisy sequences at any given timestep $t$ are calculated using a one-step Markov transition as:
\begin{equation}
\small
\mathbf{x}_t=\sqrt{\bar{\alpha}_t} \mathbf{x}_0+\sqrt{1-\bar{\alpha}_t} \epsilon
\end{equation}

where $\epsilon \sim \mathcal{N}(\mathbf{0}, \mathbf{I})$ denotes the noise sampled from Gaussian distribution.

\textbf{Boundary Indices Reverse Diffusion.}
Starting from a noise-perturbed state, the reverse diffusion process employs the non-Markovian denoising strategy DDIM \cite{shen-etal-2023-diffusionner,song2021denoising}. DDIM is for precise reconstruction of term boundaries. The process involves selecting a subsequence $\tau$ from the full timestep sequence $[1, \ldots, T]$, with a length of $\gamma$. We iteratively refining the boundary sequences $\mathbf{x}_{\tau_i}$ using the information from the preceding timestep. The iterative refinement process, utilizing a trainable denoising network $f_\theta$ conditioned on $S$ at $\tau_i$, as follows:
\begin{equation}
\small
\begin{gathered}
\hat{\mathbf{x}}_0=f_\theta\left(\mathbf{x}_{\tau_i}, S, \tau_i\right); \quad
\hat{\epsilon}_{\tau_i}=\frac{\mathbf{x}_{\tau_i}-\sqrt{\alpha_{\tau_i}} \hat{\mathbf{x}}_0}{\sqrt{1-\alpha_{\tau_i}}}
\end{gathered}
\end{equation}
where $\hat{\mathbf{x}}_0$ denotes the predicted boundary at timestep $\tau_i$, and $\hat{\epsilon}_{\tau_i}$ denotes the estimated noise. This noise is determined as the normalized difference between the perturbed boundary sequences $\mathbf{x}_{\tau_i}$ and the predicted boundary sequences $\hat{\mathbf{x}}_0$. The refined predictions are then combined with the estimated noise, adjusted by their respective standard deviations. This process is iteratively repeated, as encapsulated in the expression, $\mathbf{x}_{\tau_{i-1}}=\sqrt{\alpha_{\tau_{i-1}}} \hat{\mathbf{x}}_0+\sqrt{1-\alpha_{\tau_{i-1}}} \hat{\epsilon}_{\tau_i}$. Following $\gamma$ iterations of the DDIM, the perturbed boundary indices undergo a gradual refinement, converging towards accurate boundary indices. 

\subsection{Network Architecture} \label{sec 3.2}

Our denoising network $f_\theta\left(\mathbf{x}_{t}, S, t_i\right)$ is designed to take perturbed boundary sequences $\mathbf{x}_{t}$ and the sentence $S$ as input, and subsequently predict the corresponding term boundary $\hat{\mathbf{x}}_0$ along with the sentiment polarity. The architecture of this network, illustrated in Figure \ref{fig:model overview}, comprises two key components: a sentence encoder and a boundary indices decoder.

\textbf{Sentence Encoder.}
The Sentence Encoder transforms the input context $S=\{w_1,w_2,...,w_M\}$, with a length of $M$, into a $h$-dimensional sentence representation $\mathbf{H}_S=\{h_1,h_2,...,h_M\} \in \mathbb{R}^{M \times h}$. Our implementation involves leveraging pre-trained language models (PLMs) \cite{devlin-etal-2019-bert} with a bi-directional LSTM:
\begin{equation}
\small
\mathbf{H}_S=\operatorname{Encoder}(S)
\end{equation}

\textbf{Boundary Indices Decoder.}
The boundary indices decoder processes the sentence representation $\mathbf{H}_S$ to derive semantic representations for the corrupted sequence of boundary indices $\mathbf{x}_t$, which denote aspect and opinion terms. Initially, the noisy sequences are discretized into word indices through rescaling. The sequence representation $\mathbf{H}_X=\{h_i^X\}_{i=1}^{N_{train}} \in \mathbb{R}^{N_{train} \times h}$ is then computed by mean-pooling over the tokens at the designated start and end indices of aspect and opinion terms. Each $h_i^X$ represents the pooled representation of the $i$-th sequence within boundary sequences, calculated as follows:
\begin{equation}
\small
h_i^X=Pooling(h_{a_i^s},h_{a_i^e},h_{o_i^s},h_{o_i^e})
\end{equation}

To further refine sequence representations, we utilize a transformer decoder \cite{vaswani2017attention} integrated with self-attention and cross-attention layers. The self-attention module enhances interactions among sequences by utilizing query, key, and values derived from the sequence representations $\mathbf{H}_X$:
\begin{equation}
\small
    \mathbf{H}_{sa}=\operatorname{SelfAttention}(\mathbf{H}_X)
\end{equation}
where $\mathbf{H}_{sa}\in \mathbb{R}^{N_{train} \times h}$. In tandem, the cross-attention mechanism further refines the sequence representation by incorporating the broader semantic context of the sentence. This is achieved by utilizing the output of the self-attention module $\mathbf{H}_{sa}$ as a query, with the key and value derived from the sentence representation $\mathbf{H}_S$: 
\begin{equation}
\small
\mathbf{H}_{ca}=\operatorname{CrossAttention}(\mathbf{H}_{sa}, \mathbf{H}_S)
\end{equation}
where $\mathbf{H}_{ca}\in \mathbb{R}^{N_{train} \times h}$. To accommodate the iterative nature of the diffusion process, sinusoidal embeddings $\mathbf{E}_t$ corresponding to each timestep $t$ are integrated into the sequence representations. The final noisy sequence representations $\overline{\mathbf{H}}_X$ are calculated as follows:
\begin{equation}
\small
\overline{\mathbf{H}}_X=\mathbf{H}_{ca}+\mathbf{E}_t
\normalfont
\end{equation}

Moreover, we employ 4 index pointers to predict boundary indices of aspect and opinion terms, respectively. For each index $\delta \in \{a^s, a^e, o^s, o^e\}$, we create a fused representation $\mathbf{H}_{S X}^\delta \in \mathbb{R}^{N_{train} \times M \times h}$, which combines the noisy sequence representation with the sentence representation. The likelihood $\mathbf{P}^\delta \in \mathbb{R}^{N_{train} \times M}$ of each index being a boundary of term is as follows:
\begin{equation}
\small
\mathbf{H}_{S X}^\delta=\mathbf{W}_S^\delta \mathbf{H}_S + \mathbf{W}_X^\delta\overline{\mathbf{H}}_X; \quad
\mathbf{P}^\delta=FFN\left(\mathbf{H}_{S X}^\delta+\mathbf{E}_p^\delta\right)
\end{equation}

where $\mathbf{W}_S^\delta, \mathbf{W}_X^\delta \in \mathbb{R}^{h \times h}$ are learnable matrices, and $\operatorname{\textit{FFN}(\cdot)}$ denotes a feed-forward network (FFN). $\mathbf{E}_p^\delta \in \mathbb{R}^{N_{train} \times M \times h}$ represents type embeddings that distinguish between aspect and opinion boundaries.

\textbf{Sentiment Classifier.}
The sentiment classifier processes the sequence representations $\overline{\mathbf{H}}_X$ through a FFN to output a probability distribution: 
\begin{equation}
\small
\mathbf{P}^c=FFN\left(\overline{\mathbf{H}}_X\right)    
\end{equation}
where $\mathbf{P}^c \in \mathbb{R}^{N_{train} \times C}$, and $C$ represents the total number of sentiment polarity categories.

\textbf{Contrastive Denoising Training.}
In DiffuSent, the diffusion process introduces uncertainty, which can lead to duplicate predictions around the ground-truth boundary indices \cite{kou2023bayesdiff,du2024diffusion}. This uncertainty allows the model to explore multiple plausible start and end positions, which is beneficial for multi-word terms, but it can also cause inaccurate boundary predictions under subtle variations. To improve boundary localization and reduce false triplet generation in sentiment classification, we introduce a contrastive denoising training strategy. As shown in Figure \ref{fig:model overview}, we construct positive and negative samples by adding two noise scales, $\lambda_1$ and $\lambda_2$, to the $N_{train}$ gold boundary sequences, where $\lambda_1 < \lambda_2$. During reverse diffusion, the decoder takes both types of samples as additional inputs. Positive samples, corrupted with noise smaller than $\lambda_1$, are expected to reconstruct the corresponding ground truth. Negative samples, corrupted with noise larger than $\lambda_1$ but smaller than $\lambda_2$, are expected to be classified as ``{\tt Invalid}'', denoted by $\varepsilon$. For each ground-truth boundary sequence, we generate one positive and one negative sample, yielding $2 \times N_{train}$ samples per sentence. This process produces the boundary probabilities $\overline{\mathbf{P}}^\delta$ for positive samples, and the classification probabilities $\overline{\mathbf{P}}^c$ and $\tilde{\mathbf{P}}^c$ for positive and negative samples, respectively.

\subsection{Training Loss} \label{sec 3.3}

\textbf{Matching Loss.}
To optimally align the $N_{train}$ predictions and their corresponding $N_{train}$ expanded ground-truth values, we employ the Hungarian algorithm \cite{kuhn1955hungarian} to establish an optimal matching $\hat{\psi}$ between the two sets. In this context, $\hat{\psi}(i)$ denotes the ground-truth corresponding to the $i$-th noisy sequence. The reverse process is trained by maximizing the likelihood of the prediction:
\begin{equation}
\small
\begin{aligned}
    \mathcal{L}_m = -\sum_{i=1}^{N_{train}} \bigg[ & \sum_{\delta \in\{a^s, a^e, o^s, o^e\}} \log \mathbf{P}_i^\delta\left(\hat{\psi}^\delta(i)\right) + \log \mathbf{P}_i^c\left(\hat{\psi}^c(i)\right) \bigg]
\end{aligned}
\end{equation}

\textbf{Contrastive Denoising Loss.}
The contrastive loss consists of boundary loss and sentiment classification loss. Specifically, the boundary loss is only calculated according to boundary probabilities $\overline{\mathbf{P}}^\delta$ of positive samples. The classification loss is calculated according to classification probabilities $\overline{\mathbf{P}}^c$ and $\tilde{\mathbf{P}}^c$ for positive and negative samples, respectively.The contrastive loss is computed as:
\begin{equation}
\small
\begin{aligned}
\mathcal{L}_c=-\sum_{i=1}^{N_{train}} (&\sum_{\delta \in\{a^s, a^e, o^s, o^e\}} \log \overline{\mathbf{P}}_i^\delta\left(\hat{Y}_i^\delta\right) + \log \overline{\mathbf{P}}_i^c\left(\hat{Y}_i^c\right)+ \log \tilde{\mathbf{P}}_i^c\left(\varepsilon\right))
\end{aligned}
\normalfont
\end{equation}

We jointly optimize matching loss $\mathcal{L}_m$ and contrastive denoising loss $\mathcal{L}_c$. The overall training loss is denoted as:
\begin{equation}
\small
\mathcal{L}=\mathcal{L}_m+\mathcal{L}_c
\end{equation}



\begin{table*}[t]
\fontsize{6pt}{10pt} \selectfont 
\setlength{\tabcolsep}{0.5pt} 
\centering
\caption{Statistics of the four datasets used in our experiments.}
\begin{tabular}{@{}ccccccccccccccccccl@{}}
\toprule
\multicolumn{2}{c}{\multirow{2}{*}{Datasets}} & \multicolumn{4}{c}{Lap14} & \multicolumn{4}{c}{Res14} & \multicolumn{4}{c}{Res15} & \multicolumn{4}{c}{Res16} & \multicolumn{1}{c}{\multirow{2}{*}{Tasks}}        \\
\multicolumn{2}{c}{}   & \#s  & \#a  & \#o  & \#p  & \#s  & \#a  & \#o  & \#p  & \#s  & \#a  & \#o  & \#p  & \#s   & \#a  & \#o & \#p  & \multicolumn{1}{c}{}   \\ \midrule
\multirow{2}{*}{$D_{17}$} & train & 3048 & 2373 & 2504 & -    & 3044 & 3699 & 3484 & -    & 1315 & 1199 & 1210 & -    & -     & -    & -   & -    & \multirow{2}{*}{\begin{tabular}[c]{@{}l@{}}AE,OE,\\ ALSC\end{tabular}}        \\
  & test & 800  & 654  & 674  & -    & 800  & 1134 & 1008 & -    & 685  & 542  & 510  & -    & -     & -    & -   & -    & \\ \midrule
\multirow{2}{*}{$D_{19}$} & train & 1158 & 1634 & -    & -    & 1627 & 2643 & -    & -    & 754  & 1076 & -    & -    & 1079  & 1512 & -   & -    & \multirow{2}{*}{AOE}   \\
  & test & 343  & 482  & -    & -    & 500  & 865  & -    & -    & 325  & 436  & -    & -    & 329   & 457  & -   & -    & \\ \midrule
\multirow{3}{*}{$D_{20a}$} & train & 920  & -    & -    & 1265 & 1300 & -    & -    & 2145 & 593  & -    & -    & 923  & 842   & -    & -   & 1289 & \multirow{3}{*}{\begin{tabular}[c]{@{}l@{}}AESC,\\ AOPE,\\ ASTE\end{tabular}} \\
  & dev  & 228  & -    & -    & 337  & 323  & -    & -    & 524  & 148  & -    & -    & 238  & 210   & -    & -   & 316  & \\
  & test & 339  & -    & -    & 490  & 496  & -    & -    & 862  & 318  & -    & -    & 455  & 320   & -    & -   & 465  & \\ \midrule
\multirow{3}{*}{$D_{20b}$} & train & 906  & -    & -    & 1460 & 1266 & -    & -    & 2338 & 605  & -    & -    & 1013 & 857   & -    & -   & 1394 & \multirow{3}{*}{ASTE}  \\
  & dev  & 219  & -    & -    & 346  & 310  & -    & -    & 577  & 148  & -    & -    & 249  & 210   & -    & -   & 339  & \\
  & test & 328  & -    & -    & 543  & 492  & -    & -    & 994  & 322  & -    & -    & 485  & 326   & -    & -   & 514  & \\ \bottomrule
\end{tabular}
\label{tab:dataset}
\end{table*}

\section{Experiment}
\subsection{Experimental Setup}
\textbf{Datasets.}
We evaluate our methods across seven subtasks using four datasets from SemEval Challenges. The $D_{17}$ dataset, annotated by \cite{wang2017coupled}, comprises unpaired opinion terms, while the $D_{19}$ dataset, annotated by \cite{fan-etal-2019-target}, pairs opinion terms with corresponding aspects. Annotated by \cite{peng2020knowing}, the $D_{20a}$ dataset includes aspect labels, corresponding opinion labels, and sentiment polarities. Additionally, the $D_{20b}$ dataset, refined by \cite{xu-etal-2020-position}, eliminates triples with inaccurate sentiments and labels missing triples. We present their statistics in Table \ref{tab:dataset}.

\textbf{Baselines.}
The baselines for evaluating DiffuSent across various datasets are categorized into three groups:
\begin{itemize}[leftmargin=*]
    \item For AE, OE and ALSC subtasks on $D_{17}$, and AOE subtasks on $D_{19}$, the models include: BART-GEN \cite{yan-etal-2021-unified}, SyMux \cite{fei2022inheriting}, SK2 \cite{li2022sk2}, MvP \cite{gou-etal-2023-mvp}.
    
    \item For AESC, AOPE, ASTE subtasks on $D_{20a}$, the models are Peng-two-stage \cite{peng2020knowing}, Dual-MRC \cite{mao2021joint}, BART-GEN \cite{yan-etal-2021-unified}, LEGO-ABSA \cite{gao2022lego}, SyMux \cite{fei2022inheriting}, SK2 \cite{li2022sk2}, MvP \cite{gou-etal-2023-mvp}, PT-GCN \cite{peng2024prompt}, SAAG \cite{sun2025senticnet}, LLaMA3-8B\footnote{https://huggingface.co/meta-llama/Meta-Llama-3-8B}, GPT-4o\footnote{https://openai.com/index/hello-gpt-4o/}. 
    
    \item For ASTE subtask on $D_{20b}$, the baselines are BART-GEN \cite{yan-etal-2021-unified}, Span-ASTE \cite{xu2021learning}, UIE \cite{lu-etal-2022-unified}, SK2 \cite{li2022sk2}, SBN \cite{chen2022span}, STAGE \cite{liang2023stage}, SimSTAR \cite{li2023simple}, SLGM \cite{zhou-qian-2023-strength}, MvP \cite{gou-etal-2023-mvp}, SATPC \cite{li2024improving}, SSGCN \cite{zhang2025aspect}, QAIE \cite{lu2025qaie}, DiffuSyn \cite{yi2025diffusyn}.
\end{itemize}

\begin{table*}[t]
\fontsize{5.8pt}{8pt} \selectfont 
\setlength{\tabcolsep}{0.1pt} 
\centering
\caption{Comparison F1-scores(\%) for AESC, AOPE and ASTE on $D_{20a}$ dataset. The best and the second best F1-scores are in \textbf{bold} and \underline{underlined}, respectively. $^\dagger$ denotes the reproduced result using the released code. Results marked with $^\spadesuit$ results are retrieved from \cite{han2023information}. Results marked with "*" indicate a statistically significant improvement with $p \textless 0.01$ under the bootstrap paired t-test.}
\begin{tabular}{@{}c|c|cccccccccccc@{}}
\toprule
\multirow{2}{*}{Model} & \multirow{2}{*}{PLM} & \multicolumn{3}{c}{Lap14} & \multicolumn{3}{c}{Res14} & \multicolumn{3}{c}{Res15}   & \multicolumn{3}{c}{Res16}  \\ \cmidrule(l){3-14} 
 &        & AESC  & AOPE  & ASTE  & AESC  & AOPE  & ASTE  & AESC    & AOPE  & ASTE  & AESC    & AOPE & ASTE  \\ \midrule
Dual-MRC & Bert-base   & 64.59 & 63.37 & 55.58 & 76.57 & 74.93 & 70.32 & 65.14   & 64.97 & 57.21 & 70.84   & 75.71 & 67.40 \\
SyMux   & Roberta & 70.32        & 67.64        & 60.11 & 78.68        & \underline{79.42}        & \underline{74.84} & 69.08 & 69.82        & 63.13 & \underline{77.95} & 78.82       & 72.76 \\
SK2 & Bert-large  & 69.42 & 68.12 & 60.14 & 78.72 & 78.19 & 73.32 & 73.30   & 72.05 & 64.32 & 77.78   & 79.89 & 72.03 \\
PT-GCN & T5-base & -  & -  & \underline{62.80}  & -  & -  & 74.22  & -    & -  & 67.04  & -    & - & \underline{74.68}  \\
SAAG & Bert-base & -  & -  & 59.58  & -  & -  & 73.76  & -    & -  & 63.48  & -    & - & 71.01  \\\midrule
BART-GEN & Bart-base   & 68.17 & 66.11 & 57.59 & 78.47 & 77.68 & 72.46 & 69.95   & 67.98 & 60.11 & 75.69   & 77.38 & 69.98 \\
LEGO-ABSA & T5-base & \underline{72.30}  & 71.30  & 62.20  & \underline{80.60}  & 78.10  & 73.70  & 74.20    & 72.90  & 64.40  & 76.10    & 77.60 & 71.50  \\
MvP$^\dagger$  & T5-base & 70.55 & \underline{71.38} & 62.42 & 78.06 & 77.95 & 74.60 & \underline{74.84}       & \underline{74.06} &65.25 &77.63       & \underline{80.46}    &73.28 \\
 ChatGPT-4$^\spadesuit$ & -   & 52.70 & 42.80 & - & 65.93 & 59.70 & - & 65.71   & 55.52 & - & 63.38   & 57.93 & - \\
  ChatGPT-4o$^\spadesuit$  & -   & - & - & 44.63 & - & - & 63.57 & -   & - & 56.53 & -   & - & 64.68 \\
 LLaMA3-8B$^\spadesuit$  & -   & - & - & 36.40 & - & - & 53.49 & -   & - & 46.73 & -   & - & 56.01 \\
 \midrule
DiffuSent & Bert-base   & \textbf{73.74$^*$} & \textbf{71.67$^*$} & \textbf{63.31$^*$} & \textbf{81.13$^*$} & \textbf{79.86$^*$} & \textbf{74.91$^*$} & \textbf{75.85$^*$}        & \textbf{74.19$^*$} & \textbf{67.58$^*$} & \textbf{79.16$^*$}        & \textbf{80.90$^*$} & \textbf{75.09$^*$} \\\bottomrule
\end{tabular}
\label{tab:mainresult}
\end{table*}

\begin{figure}[t]
    \centering
    \includegraphics[width=0.75\textwidth]{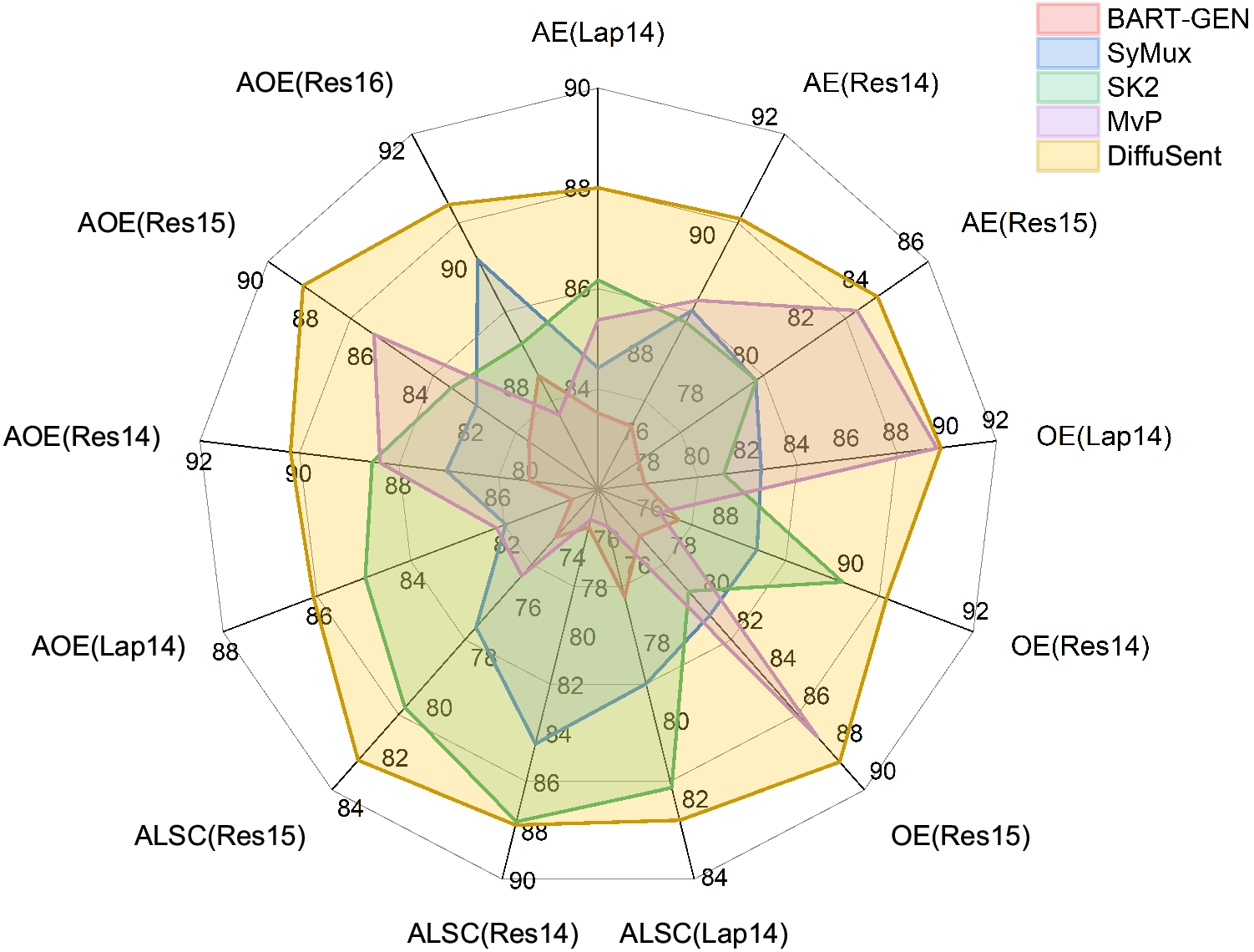}
    \caption{Comparison F1-scores for AE, OE, ALSC on the $D_{17}$ dataset, and AOE on the $D_{19}$ dataset.}
    \label{fig:appendix_radar}
\end{figure}

\begin{table}[t]
\fontsize{7pt}{8pt} \selectfont 
\setlength{\tabcolsep}{15pt} 
\centering
\caption{Comparison F1-scores(\%) for ASTE on $D_{20b}$ dataset. Symbols have the same meanings as in Table \ref{tab:mainresult}.}
\begin{tabular}{@{}cccccc@{}}
\toprule
Model     & PLM        & Lap14 & Res14 & Res15 & Res16 \\ \midrule
Span-ASTE & Bert-base  & 59.38 & 71.85 & 63.27 & 70.26 \\
SK2  & Bert-large & 60.56 & 73.27 & 65.00 & 72.19 \\
SBN & Bert-base  & 62.65 & 74.34    & 64.82 & 72.08 \\
SimSTAR$^\dagger$  & Bert-base  & 59.98 & 70.15 & 63.5 & 70.25 \\
STAGE$^\dagger$ & Bert-base  & 59.58 & 72.58 & 63.49 & 71.06 \\
SATPC & Bert-base    & 62.59 & \underline{74.79} & 65.03 & 71.17 \\
SSGCN  & Bert-base    & 59.74 & 73.05 & 61.74 & 68.57 \\
\midrule
BART-GEN & Bart-base  & 58.69 & 65.25 & 59.26 & 67.62 \\
UIE-base & T5-base    & 62.94 & 72.55 & 64.41 & 72.86 \\
ChatGPT-4$^\spadesuit$  & -  & 39.01 & 54.89 & 47.88 & 56.55 \\
MvP$^\dagger$ & T5-base    & 61.51 & 73.48 & 64.65    & 73.38    
\\
SLGM$^\dagger$ & T5-base    & \textbf{63.28} & 73.39 & \underline{65.72} & \underline{73.41} \\
QAIE & T5-base    & 38.58 & - & 43.82 & 51.41 \\
DiffuSyn & Bert-large    & 61.41 & 73.74 & 64.55 & 72.48 \\
\midrule
DiffuSent & Bert-base  & \underline{63.03$^*$}    & \textbf{74.97$^*$} & \textbf{66.39$^*$} & \textbf{74.22$^*$} \\ \bottomrule
\end{tabular}
\label{tab:mainresult2}
\end{table}

\subsection{Main Results}
We adopt F1-score as the evaluation metric for all experiments. A predicted ABSA tuple is regarded as correct only under exact match with the gold tuple in all its elements. We make the following observations.
\begin{itemize}[leftmargin=*]
    \item \textbf{DiffuSent delivers consistent improvements across AESC, AOPE, and ASTE subtasks on $D_{20a}$.} As shown in Table \ref{tab:mainresult}, DiffuSent achieves state-of-the-art performance in all twelve settings. Relative to the strongest unified baselines per column, DiffuSent achieves improvements between $0.07$ and $1.21$ F1-scores, with an average gain of roughly $0.6$. All reported improvements are statistically significant under a paired bootstrap test at $p < 0.01$. These results highlight that progressive boundary denoising effectively sharpens aspect–opinion span localization under the exact-match criterion.
    \item \textbf{Superior ASTE Performance on $D_{20b}$ Across Most Domains.} 
In comparison to the latest ASTE benchmarks, as shown in Table \ref{tab:mainresult2}, DiffuSent demonstrates superior performance. Even when compared against auto-regressive generative systems including UIE, MvP, and SLGM, all of which employ the substantially larger \textit{T5-base} backbone, DiffuSent still delivers improvements of +0.94, +0.67, and +0.81 F1 on {\tt Res14}, {\tt Res15}, and {\tt Res16}, respectively. These findings confirm that our performance gains originate from the diffusion-based boundary refinement rather than from scaling pre-trained backbones.
    \item \textbf{Large Language Models Exhibit Systematic Weaknesses in ABSA tasks.}
As reflected in Tables~\ref{tab:mainresult} and~\ref{tab:mainresult2}, ChatGPT-4 and ChatGPT-4o fall significantly short on all ABSA subtasks, often trailing even earlier BERT-based architectures by a considerable margin. The gap becomes especially pronounced in extraction-heavy settings such as AOPE and ASTE, where precise boundary identification and strict tuple consistency are required. 

    \item \textbf{Consistent Advantages Across Extraction and Classification Tasks on $D_{17}$ and $D_{19}$.}
Figure~\ref{fig:appendix_radar} shows that DiffuSent consistently forms the outer envelope relative to representative unified baselines across AE, OE, and ALSC on $D_{17}$ and AOE on $D_{19}$. This uniform dominance across both extraction-oriented tasks (AE, OE, AOE) and opinion classification (ALSC) illustrates that the underlying diffusion mechanism enhances not only boundary precision but also downstream semantic consistency. 

\end{itemize}

\subsection{Ablation Study}
To further investigate the impact of each component and hyper-parameter in DiffuSent, we conduct comprehensive ablation studies on ASTE task on {\tt Res15} and {\tt Res16} from $D_{20b}$ in Table \ref{tab:ablation}.

\textbf{Contrastive Denoising. }
Removing the contrastive denoising (CD) objective leads to clear degradation on both datasets, with F1-score drops of 2.23 and 2.78 points on {\tt Res15} and {\tt Res16}, respectively. The decline reflects the challenge posed by subtle boundary variations introduced during the diffusion process. without CD, the model more frequently produces near-duplicate spans or spurious triplets. The contrastive objective effectively suppresses these ambiguous alternatives by enforcing discriminability between clean and corrupted representations, thereby stabilizing boundary localization.

\begin{table}[t]
\renewcommand{\arraystretch}{0.7}
\centering
\fontsize{8pt}{10pt} \selectfont 
\setlength{\tabcolsep}{20pt} 
\caption{Ablation (F1-score,\%) on {\tt Res15} and {\tt Res16} on ASTE subtasks. The best results are marked in \textbf{bold}.}
\begin{tabular}{@{}cccc@{}}
\toprule
\multicolumn{2}{c}{Settings}     & Res15 & Res16 \\ \midrule
\multirow{2}{*}{\begin{tabular}[c]{@{}c@{}}Contrastive\\ Denoising\end{tabular}} & \ding{55}    & 64.16 & 71.44 \\ &  \ding{51}    & \textbf{66.39} & \textbf{74.22} \\ \midrule
\multirow{6}{*}{\begin{tabular}[c]{@{}c@{}}Duffusion\\ Timestep\end{tabular}}
& 700 & 63.55 & 70.18 \\ & 800 & 64.03 & 72.51 \\ & 900 & 65.41 & 72.88 \\ & 1000 & \textbf{66.39} & \textbf{74.22} \\ & 1500 & 64.42 & 71.40  \\ & 2000 &65.57       & 71.22 \\ \midrule \multirow{3}{*}{\begin{tabular}[c]{@{}c@{}}Number of\\ Noisy Sequence\end{tabular}} & 30   & 63.53 & 72.23 \\& 60   & \textbf{66.39} & \textbf{74.22} \\& 90   & 64.61 & 72.26 \\ \bottomrule
\end{tabular}
\label{tab:ablation}
\end{table}

\textbf{Diffusion Timestep. }
The diffusion timestep determines the magnitude of perturbation applied in the forward process. Table~\ref{tab:ablation} shows that deviating from the default setting of 1000 timesteps in either direction harms performance. Smaller timesteps (700–900) fail to introduce sufficient variation for effective denoising refinement, while larger values (1500–2000) inject excessive noise that obscures true boundary cues. This finding underscores that DiffuSent benefits from a moderate level of perturbation, where the injected noise is large enough to encourage robust boundary reconstruction but not so large that it overwhelms useful semantic structure.

\textbf{Number of Noisy Sequence. }
We also vary the number of noisy sequences used during training and inference. The best performance is achieved with 60 sequences, whereas both smaller (30) and larger (90) values lead to a reduction in F1. Using too few sequences limits the model's exposure to uncertainty, reducing its ability to generalize across boundary perturbations. Conversely, using too many sequences increases the tendency to generate redundant or varied predictions, complicating the identification of correct triplets.

\begin{figure}[t]
\centering
	\subfloat[single-word triplets]{
		\includegraphics[scale=0.22]{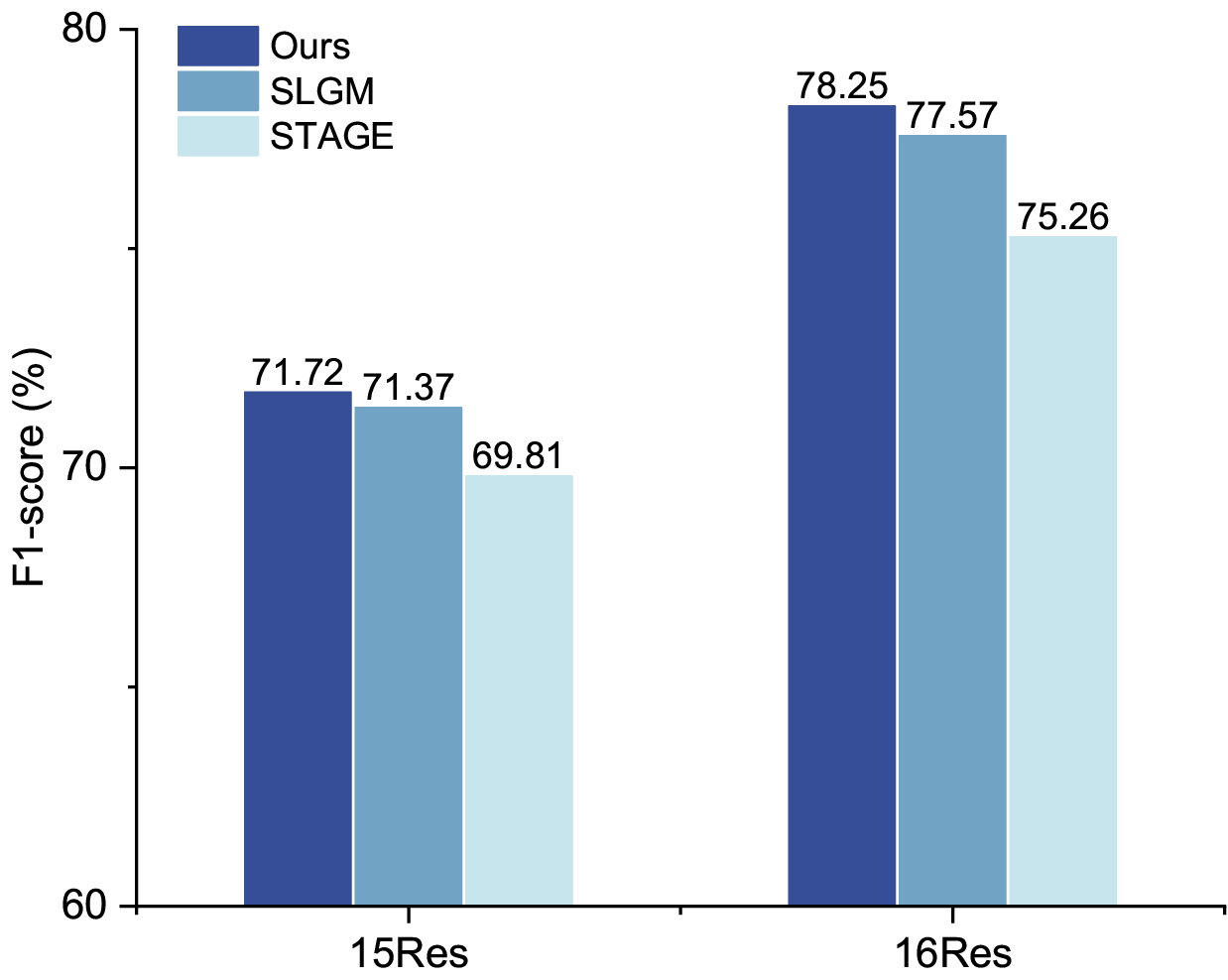}
	}%
	\subfloat[multi-word triplets]{
		\includegraphics[scale=0.22]{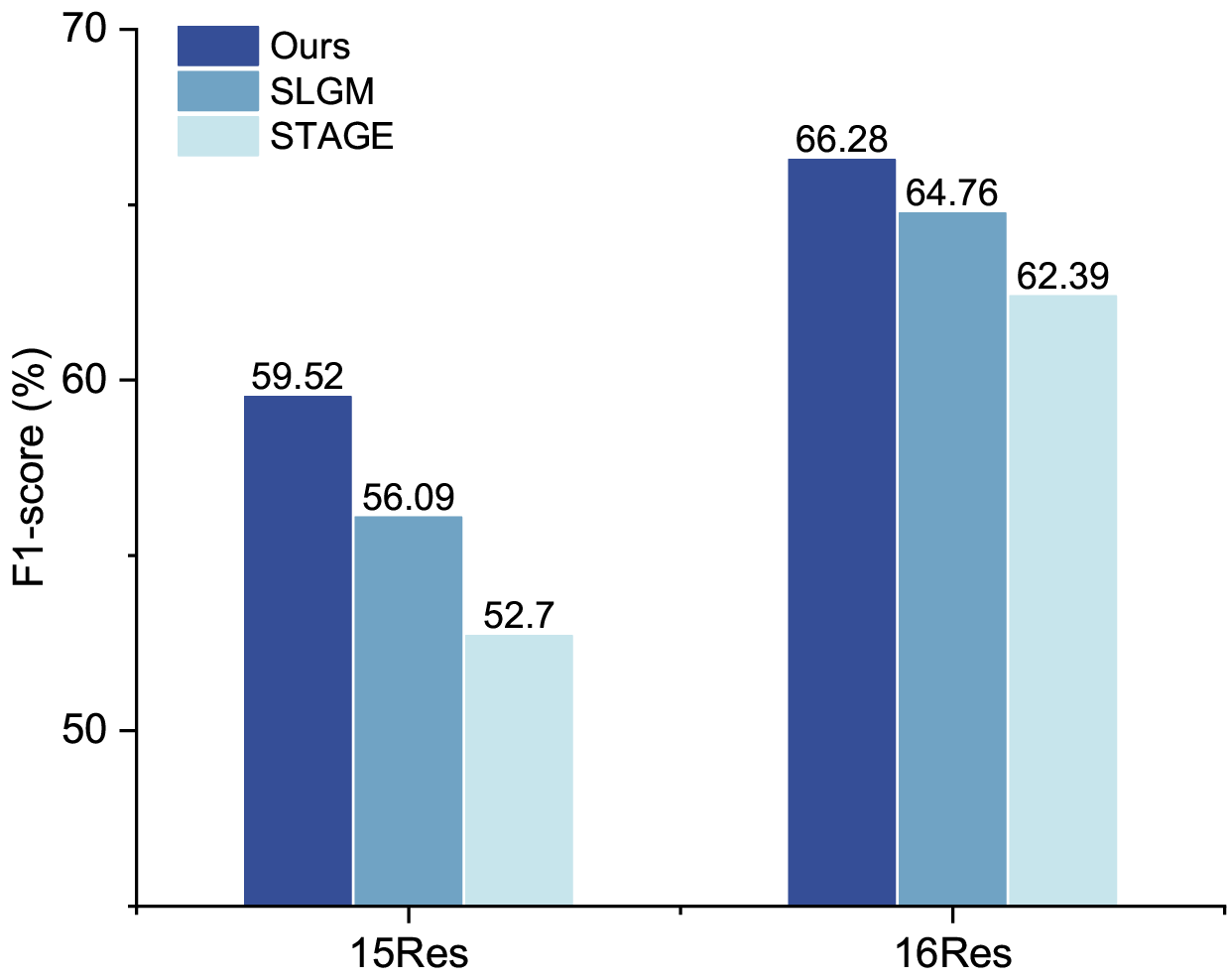}
	}%
	\centering
\label{single-result}
\caption{F1-scores of DiffuSent on multi-word and single-word triplets compared with SLGM and STAGE.} 
\label{multi and single result}
\end{figure}

\subsection{Performance on Multi-word Triplets}

According to statistic data \cite{zhou-qian-2023-strength}, multi-word triplets account for roughly one-third of all triplets. To assess DiffuSent's capability with multi-word terms, we focus on triplets containing at least one multi-word aspect or opinion term, contrasting it with single-word triplets. Our evaluation includes comparisons with the latest span-based approach, STAGE \cite{liang2023stage}, and a generative method, SLGM \cite{zhou-qian-2023-strength}, on the {\tt Res15} and {\tt Res16} datasets from $D_{20b}$. As shown in Figure \ref{multi and single result}, our model consistently outperforms others across various metrics. Notably, DiffuSent exhibits a more substantial improvement, achieving an average F1-score increase of 2.48\% for multi-word triplets compared to a 0.52\% increase for single-word triplets. These results underscore DiffuSent's effectiveness in identifying the boundaries of multi-word terms, consequently enhancing the overall performance.



\begin{table*}[t]
\fontsize{7pt}{8pt} \selectfont 
\renewcommand{\arraystretch}{1.2}
\caption{Qualitative Comparison of DiffuSent with and without Contrastive Denoising (CD) on Noise and Boundary Resolution. \ding{51} indicates a correct triplet prediction, while \ding{55} denotes a wrong prediction. Note that spurious or extra triplet predictions are counted as incorrect, negatively impacting the overall F1-score.}
\resizebox{\linewidth}{!}{
\begin{tabular}{@{}l@{}}
\toprule
\textit{Test Sentence: Oh Speaking of bathroom, the mens bathroom was disgusting. } \\ \midrule
\textbf{Gold Triplet}: \textless{}mens bathroom, disgusting, negative\textgreater{} \\
\textbf{Diffusent}: \textless{}mens bathroom, disgusting, negative\textgreater{}\ding{51} \\
\textbf{Diffusent w/o CD}: \textless{}bathroom, disgusting, negative\textgreater{}\ding{51}, \textless{}mens bathroom, disgusting, negative\textgreater{}\ding{55}   \\ \midrule
\textit{Test Sentence: Paul, the mairew d', was totally professional and always on top of things.}       \\ \midrule
\textbf{Gold Triplet:} \textless{}Paul, professional, positive\textgreater{} \\
\textbf{DiffuSent:} \textless{}Paul, professional, positive\textgreater{}\ding{51}    \\
\textbf{DiffuSent w/o CD}: \textless{}Paul, professional, positive\textgreater{}\ding{51}, \textless{}maitre d', professional, positive\textgreater{}\ding{55}   \\ \midrule
\textit{Test Sentence: The service is amazing, I've had different waiters and they were all nice, which is a rare thing in NYC.}                   \\ \midrule
\textbf{Gold Triplet:} \textless{}service, amazing, positive\textgreater{}, \textless{}waiters, nice, positive\textgreater{}                       \\
\textbf{DiffuSent:} \textless{}service, amazing, positive\textgreater{}\ding{51}, \textless{}waiters, nice, positive\textgreater{}\ding{51}                          \\
\textbf{DiffuSent w/o CD:} \textless{}service, amazing, positive\textgreater{}\ding{51}, \textless{}waiters, nice, positive\textgreater{}\ding{51}, \textless{}waiters, rare, positive\textgreater{}\ding{55}                                     \\ \midrule
\textit{Test Sentence: Shame on this place for the horrible rude staff and non-existent customer service.}                                         \\ \midrule
\textbf{Gold Triplet:} \textless{}stuff, rude, negative\textgreater{}, \textless{}customer service, non-existent, negative\textgreater{}           \\
\textbf{DiffuSent:} \textless{}stuff, rude, negative\textgreater{}\ding{51}, \textless{}customer service, non-existent, negative\textgreater{}\ding{51}              \\
\textbf{DiffuSent w/o CD:} \textless{}stuff, rude, negative\textgreater{}\ding{51}, \textless{}customer service, non-existent, negative\textgreater{}\ding{51}, \textless{}stuff, shame, negative\textgreater{}\ding{55}                                          \\ \midrule
\textit{Test Sentence: Food was amazing - I love Indian food and eay it quite regularly, but I can say this is one of the best I've had.}          \\ \midrule
\textbf{Gold Triplet:} \textless{}Food, amazing, positive\textgreater{}             \\
\textbf{DiffuSent:} \textless{}Food, amazing, positive\textgreater{}\ding{51}, \textless{}Indian food, best, positive\textgreater{}\ding{55}                         \\
\textbf{DiffuSent w/o CD:} \textless{}Food, amazing, positive\textgreater{}\ding{51}, \textless{}Indian food, best, positive\textgreater{}\ding{55}, \textless{}Indian food, love, positive\textgreater{}\ding{55}, \textless{}Food, love, positive\textgreater{}\ding{55} \\ \bottomrule
\end{tabular}}
\label{CD}
\end{table*}

\subsection{Effective of Contrastive Denoising Scheme}

Table \ref{CD} provides qualitative comparisons on {\tt Res15}, illustrating how DiffuSent behaves with and without the contrastive denoising component. Across all examples, the model trained without this scheme exhibits two recurring failure modes. (i) Boundary drift induced by diffusion noise, leading to subtle but semantically consequential deviations such as predicting "bathroom" instead of the correct span "mens bathroom"; (ii) the absence of CD increases the likelihood of generating spurious triplets that do not correspond to any sentiment expression in the input sentence, thereby harming overall precision. In contrast, the full DiffuSent model reliably suppresses these erroneous alternatives and consistently recovers the correct <aspect, opinion, polarity> structure. The improvement suggests that the CD objective effectively sharpens the decision boundary between plausible and implausible span candidates by encouraging the model to contrast clean and corrupted boundary representations.



\begin{table}[t]
\centering
\fontsize{8pt}{10pt} \selectfont 
\setlength{\tabcolsep}{10pt} 
\caption{Computational efficiency with generative methods on {\tt Res16} from $D_{20b}$. All experiments are conducted on the same setting.}
\begin{tabular}{@{}cccccc@{}}
\toprule
    & Model     & Params. (Mb)    & F1-score   & Sents/s & SpeedUp \\ \midrule
\multirow{3}{*}{\rotatebox[origin=c]{90}{Training}}  & MvP       & 223M & -    & 9.24   & 1.00×   \\
    & SLGM      & 225M & -    & 41.36  & 4.48×   \\ \cmidrule(l){2-6} 
    & DiffuSent & 112M & -    & \textbf{71.8}   & \textbf{7.77×}   \\ \midrule
\multirow{5}{*}{\rotatebox[origin=c]{90}{Inference}} & MvP       & 223M & 73.38 & 0.86   & 1.00×   \\
    & SLGM      & 225M & 73.41 & 24.41  & 28.38×  \\ \cmidrule(l){2-6} 
    & DiffuSent$_{[\gamma=1]}$ & 112M & 73.9 & \textbf{155.98} & \textbf{181.37×} \\
    & DiffuSent$_{[\gamma=5]}$ & 112M & 74.22 & 92.61  & 106.98× \\
    & DiffuSent$_{[\gamma=10]}$ & 112M & \textbf{74.3} & 61.51  & 71.52×  \\ \bottomrule
\end{tabular}
\label{tab:inference_speed}
\end{table}

\subsection{Computational Efficiency}

To comprehensively evaluate the computational efficiency of DiffuSent, we compare its parameter footprint, training speed, and inference latency against state-of-the-art generative models, namely MvP and SLGM. As detailed in Table \ref{tab:inference_speed}, DiffuSent operates with approximately half the parameters (112M) of the baselines while consistently delivering superior predictive performance. During training, DiffuSent accelerates processing by 7.77× compared to MvP. More notably, during inference, DiffuSent achieves a higher F1-score across all denoising timesteps ($\gamma$) while maintaining substantially higher throughput. Even under the most computationally demanding setting ($\gamma$ = 10), DiffuSent achieves an F1-score of 74.3 and remains 71.5× and 2.5× faster than MvP and SLGM, respectively. This significant efficiency gain is primarily attributed to our non-autoregressive framework, which generates all triplets in parallel and effectively bypasses the bottleneck of step-by-step linearized sequence generation.

\section{Conclusion}
In this paper, we propose DiffuSent, a novel generative framework for unified aspect-based sentiment analysis (ABSA) that formulate all ABSA subtasks as boundary denoising diffusion process. Different from autoregressive token-by-token generation, DiffuSent explicitly models boundary indices and allows for dynamically refinements in interpreting complex linguistic structures like multi-word terms. In addition, to address duplicate predictions with subtle variations arising from diffusion process uncertainties, we design a contrastive denoising training that further refine aspect and opinion term boundaries. Experimental results demonstrate that DiffuSent yields a new state-of-the-art performance, showcasing superior performance in processing complex linguistic structures.
%
%
%
\bibliographystyle{splncs04}
\bibliography{Reference}

@inproceedings{wang2017coupled,
  title={Coupled multi-layer attentions for co-extraction of aspect and opinion terms},
  author={Wang, Wenya and Pan, Sinno Jialin and Dahlmeier, Daniel and Xiao, Xiaokui},
  booktitle={Proceedings of the AAAI conference on artificial intelligence},
  volume={31},
  number={1},
  year={2017}
}

@inproceedings{tang2015effective,
  title={Effective LSTMs for Target-Dependent Sentiment Classification},
  author={Tang, Duyu and Qin, Bing and Feng, Xiaocheng and Liu, Ting},
  booktitle={Proceedings of the 26th International Conference on Computational Linguistics},
  pages={3298--3307},
  year={2016}
}

@inproceedings{liang2023stage,
  title={STAGE: span tagging and greedy inference scheme for aspect sentiment triplet extraction},
  author={Liang, Shuo and Wei, Wei and Mao, Xian-Ling and Fu, Yuanyuan and Fang, Rui and Chen, Dangyang},
  booktitle={Proceedings of the AAAI Conference on Artificial Intelligence},
  volume={37},
  number={11},
  pages={13174--13182},
  year={2023}
}

@inproceedings{mao2021joint,
  title={A joint training dual-mrc framework for aspect based sentiment analysis},
  author={Mao, Yue and Shen, Yi and Yu, Chao and Cai, Longjun},
  booktitle={Proceedings of the AAAI conference on artificial intelligence},
  volume={35},
  number={15},
  pages={13543--13551},
  year={2021}
}

@inproceedings{gao2022lego,
  title={LEGO-ABSA: A prompt-based task assemblable unified generative framework for multi-task aspect-based sentiment analysis},
  author={Gao, Tianhao and Fang, Jun and Liu, Hanyu and Liu, Zhiyuan and Liu, Chao and Liu, Pengzhang and Bao, Yongjun and Yan, Weipeng},
  booktitle={Proceedings of the 29th international conference on computational linguistics},
  pages={7002--7012},
  year={2022}
}

@inproceedings{fei2022inheriting,
  title={Inheriting the wisdom of predecessors: A multiplex cascade framework for unified aspect-based sentiment analysis},
  author={Fei, Hao and Li, Fei and Li, Chenliang and Wu, Shengqiong and Li, Jingye and Ji, Donghong},
  booktitle={Proceedings of the Thirty-First International Joint Conference on Artificial Intelligence, IJCAI},
  pages={4096--4103},
  year={2022}
}

@inproceedings{gou-etal-2023-mvp,
    title = "{M}v{P}: Multi-view Prompting Improves Aspect Sentiment Tuple Prediction",
    author = "Gou, Zhibin  and
      Guo, Qingyan  and
      Yang, Yujiu",
    booktitle = "Proceedings of the 61st Annual Meeting of the Association for Computational Linguistics",
    year = "2023",
    address = "Toronto, Canada",
    publisher = "Association for Computational Linguistics",
    pages = "4380--4397",
}

@article{hoogeboom2021argmax,
  title={Argmax flows and multinomial diffusion: Learning categorical distributions},
  author={Hoogeboom, Emiel and Nielsen, Didrik and Jaini, Priyank and Forr{\'e}, Patrick and Welling, Max},
  journal={Advances in Neural Information Processing Systems},
  volume={34},
  pages={12454--12465},
  year={2021}
}

@article{austin2021structured,
  title={Structured denoising diffusion models in discrete state-spaces},
  author={Austin, Jacob and Johnson, Daniel D and Ho, Jonathan and Tarlow, Daniel and Van Den Berg, Rianne},
  journal={Advances in Neural Information Processing Systems},
  volume={34},
  pages={17981--17993},
  year={2021}
}

@article{li2022diffusion,
  title={Diffusion-lm improves controllable text generation},
  author={Li, Xiang and Thickstun, John and Gulrajani, Ishaan and Liang, Percy S and Hashimoto, Tatsunori B},
  journal={Advances in Neural Information Processing Systems},
  volume={35},
  pages={4328--4343},
  year={2022}
}

@inproceedings{gong2023diffuseq,
  title={DiffuSeq: Sequence to Sequence Text Generation with Diffusion Models},
  author={Gong, Shansan and Li, Mukai and Feng, Jiangtao and Wu, Zhiyong and Kong, Lingpeng},
  booktitle={International Conference on Learning Representations (ICLR 2023)},
  year={2023}
}

@inproceedings{peng2020knowing,
  title={Knowing what, how and why: A near complete solution for aspect-based sentiment analysis},
  author={Peng, Haiyun and Xu, Lu and Bing, Lidong and Huang, Fei and Lu, Wei and Si, Luo},
  booktitle={Proceedings of the AAAI conference on artificial intelligence},
  volume={34},
  number={05},
  pages={8600--8607},
  year={2020}
}

@inproceedings{
song2021denoising,
title={Denoising Diffusion Implicit Models},
author={Jiaming Song and Chenlin Meng and Stefano Ermon},
booktitle={International Conference on Learning Representations},
year={2021},
}

@article{kuhn1955hungarian,
  title={The Hungarian method for the assignment problem},
  author={Kuhn, Harold W},
  journal={Naval research logistics quarterly},
  volume={2},
  number={1-2},
  pages={83--97},
  year={1955},
  publisher={Wiley Online Library}
}

@inproceedings{li2022sk2,
  title={SK2: Integrating Implicit Sentiment Knowledge and Explicit Syntax Knowledge for Aspect-Based Sentiment Analysis},
  author={Li, Jia and Zhao, Yuyuan and Jin, Zhi and Li, Ge and Shen, Tao and Tao, Zhengwei and Tao, Chongyang},
  booktitle={Proceedings of the 31st ACM International Conference on Information \& Knowledge Management},
  pages={1114--1123},
  year={2022}
}

@inproceedings{lu-etal-2022-unified,
    title = "Unified Structure Generation for Universal Information Extraction",
    author = "Lu, Yaojie  and
      Liu, Qing  and
      Dai, Dai  and
      Xiao, Xinyan  and
      Lin, Hongyu  and
      Han, Xianpei  and
      Sun, Le  and
      Wu, Hua",
    booktitle = "Proceedings of the 60th Annual Meeting of the Association for Computational Linguistics",
    year = "2022",
    address = "Dublin, Ireland",
    publisher = "Association for Computational Linguistics",
    pages = "5755--5772",
}

@inproceedings{shen-etal-2023-diffusionner,
    title = "{D}iffusion{NER}: Boundary Diffusion for Named Entity Recognition",
    author = "Shen, Yongliang  and
      Song, Kaitao  and
      Tan, Xu  and
      Li, Dongsheng  and
      Lu, Weiming  and
      Zhuang, Yueting",
    booktitle = "Proceedings of the 61st Annual Meeting of the Association for Computational Linguistics",
    month = jul,
    year = "2023",
    publisher = "Association for Computational Linguistics",
    url = "https://aclanthology.org/2023.acl-long.215",
    doi = "10.18653/v1/2023.acl-long.215",
    pages = "3875--3890",
}

@inproceedings{fan-etal-2019-target,
    title = "Target-oriented Opinion Words Extraction with Target-fused Neural Sequence Labeling",
    author = "Fan, Zhifang  and
      Wu, Zhen  and
      Dai, Xin-Yu  and
      Huang, Shujian  and
      Chen, Jiajun",
    booktitle = "Proceedings of the 2019 Conference of the North {A}merican Chapter of the Association for Computational Linguistics",
    year = "2019",
    publisher = "Association for Computational Linguistics",
    pages = "2509--2518",
}

@inproceedings{hu-etal-2019-open,
    title = "Open-Domain Targeted Sentiment Analysis via Span-Based Extraction and Classification",
    author = "Hu, Minghao  and
      Peng, Yuxing  and
      Huang, Zhen  and
      Li, Dongsheng  and
      Lv, Yiwei",
    booktitle = "Proceedings of the 57th Annual Meeting of the Association for Computational Linguistics",
    year = "2019",
    address = "Florence, Italy",
    publisher = "Association for Computational Linguistics",
    pages = "537--546",
}

@inproceedings{yan-etal-2021-unified,
    title = "A Unified Generative Framework for Aspect-based Sentiment Analysis",
    author = "Yan, Hang  and
      Dai, Junqi  and
      Ji, Tuo  and
      Qiu, Xipeng  and
      Zhang, Zheng",
    booktitle = "Proceedings of the 59th Annual Meeting of the Association for Computational Linguistics",
    year = "2021",
    pages = "2416--2429",
}

@inproceedings{xu-etal-2020-position,
    title = "Position-Aware Tagging for Aspect Sentiment Triplet Extraction",
    author = "Xu, Lu  and
      Li, Hao  and
      Lu, Wei  and
      Bing, Lidong",
    booktitle = "Proceedings of the 2020 Conference on Empirical Methods in Natural Language Processing",
    year = "2020",
    pages = "2339--2349",
}

@inproceedings{pontiki2016semeval,
  title={SemEval-2016 Task 5: Aspect Based Sentiment Analysis},
  author={Pontiki, Maria and Galanis, Dimitrios and Papageorgiou, Harris and Androutsopoulos, Ion and Manandhar, Suresh and Mohammad, AL-Smadi and Al-Ayyoub, Mahmoud and Zhao, Yanyan and Qin, Bing and De Clercq, Orphee and others},
  booktitle={Proceedings of the 10th International Workshop on Semantic Evaluation (SemEval-2016)},
  pages={19--30},
  year={2016}
}

@inproceedings{zhang-etal-2021-towards-generative,
    title = "Towards Generative Aspect-Based Sentiment Analysis",
    author = "Zhang, Wenxuan  and
      Li, Xin  and
      Deng, Yang  and
      Bing, Lidong  and
      Lam, Wai",
    booktitle = "Proceedings of the 59th Annual Meeting of the Association for Computational Linguistics)",
    address = "Online",
    publisher = "Association for Computational Linguistics",
    pages = {504--510},
    year={2021}
}

@inproceedings{ma2019exploring,
  title={Exploring sequence-to-sequence learning in aspect term extraction},
  author={Ma, Dehong and Li, Sujian and Wu, Fangzhao and Xie, Xing and Wang, Houfeng},
  booktitle={Proceedings of the 57th annual meeting of the association for computational linguistics},
  pages={3538--3547},
  year={2019}
}

@inproceedings{zhao2020spanmlt,
  title={Spanmlt: A span-based multi-task learning framework for pair-wise aspect and opinion terms extraction},
  author={Zhao, He and Huang, Longtao and Zhang, Rong and Lu, Quan and Xue, Hui},
  booktitle={Proceedings of the 58th annual meeting of the association for computational linguistics},
  pages={3239--3248},
  year={2020}
}

@inproceedings{tang2015document,
  title={Document modeling with gated recurrent neural network for sentiment classification},
  author={Tang, Duyu and Qin, Bing and Liu, Ting},
  booktitle={Proceedings of the 2015 conference on empirical methods in natural language processing},
  pages={1422--1432},
  year={2015}
}

@inproceedings{liu2023Aspect,
  title={Aspect-oriented Opinion Alignment Network for Aspect-Based Sentiment Classification},  
  author={Liu, Xueyi and Hou, Rui and Gan, Yanglei and Luo, Da and Li, Changlin and Shi, Xiaojun and Liu, Qiao},  
  booktitle={Proceedings of the 26th European Conference on Artificial Intelligence},  
  pages={1552--1559},  
  year={2023}  
}

@inproceedings{zhang2022boundary,
  title={Boundary-driven table-filling for aspect sentiment triplet extraction},
  author={Zhang, Yice and Yang, Yifan and Li, Yihui and Liang, Bin and Chen, Shiwei and Dang, Yixue and Yang, Min and Xu, Ruifeng},
  booktitle={Proceedings of the 2022 Conference on Empirical Methods in Natural Language Processing},
  pages={6485--6498},
  year={2022}
}

@inproceedings{jing2021seeking,
    title = "Seeking Common but Distinguishing Difference, A Joint Aspect-based Sentiment Analysis Model",
    author = "Jing, Hongjiang  and
      Li, Zuchao  and
      Zhao, Hai  and
      Jiang, Shu",
    booktitle = "Proceedings of the 2021 Conference on Empirical Methods in Natural Language Processing",
    year = "2021",
    publisher = "Association for Computational Linguistics",
    pages = "3910--3922",
}

@inproceedings{xu2021learning,
  title={Learning Span-Level Interactions for Aspect Sentiment Triplet Extraction},
  author={Xu, Lu and Chia, Yew Ken and Bing, Lidong},
  booktitle={Proceedings of the 59th Annual Meeting of the Association for Computational Linguistics and the 11th International Joint Conference on Natural Language Processing},
  pages={4755--4766},
  year={2021}
}

@inproceedings{mukherjee2021paste,
  title={PASTE: A Tagging-Free Decoding Framework Using Pointer Networks for Aspect Sentiment Triplet Extraction},
  author={Mukherjee, Rajdeep and Nayak, Tapas and Butala, Yash and Bhattacharya, Sourangshu and Goyal, Pawan},
  booktitle={Proceedings of the Conference on Empirical Methods in Natural Language Processing},
  pages={9279--9291},
  year={2021}
}

@inproceedings{chen2022span,
  title={A span-level bidirectional network for aspect sentiment triplet extraction},
  author={Chen, Yuqi and Keming, Chen and Sun, Xian and Zhang, Zequn},
  booktitle={Proceedings of the 2022 Conference on Empirical Methods in Natural Language Processing},
  pages={4300--4309},
  year={2022}
}

@inproceedings{zhou-qian-2023-strength,
    title = "On the Strength of Sequence Labeling and Generative Models for Aspect Sentiment Triplet Extraction",
    author = "Zhou, Shen  and
      Qian, Tieyun",
    booktitle = "Findings of the Association for Computational Linguistics",
    year = "2023",
    address = "Toronto, Canada",
    publisher = "Association for Computational Linguistics",
    pages = "12038--12050",
}

@article{xiao2023survey,
  title={A survey on non-autoregressive generation for neural machine translation and beyond},
  author={Xiao, Yisheng and Wu, Lijun and Guo, Junliang and Li, Juntao and Zhang, Min and Qin, Tao and Liu, Tie-yan},
  journal={IEEE Transactions on Pattern Analysis and Machine Intelligence},
  year={2023},
  publisher={IEEE}
}

@ARTICLE{Fei21Nonautoregressive,
  author={Fei, Hao and Ren, Yafeng and Zhang, Yue and Ji, Donghong},
  journal={IEEE Transactions on Neural Networks and Learning Systems}, 
  title={Nonautoregressive Encoder–Decoder Neural Framework for End-to-End Aspect-Based Sentiment Triplet Extraction}, 
  year={2023},
  volume={34},
  number={9},
  pages={5544-5556}
}

@InProceedings{pmlr-v37-sohl-dickstein15,
  title = 	 {Deep Unsupervised Learning using Nonequilibrium Thermodynamics},
  author = 	 {Sohl-Dickstein, Jascha and Weiss, Eric and Maheswaranathan, Niru and Ganguli, Surya},
  booktitle = 	 {Proceedings of the 32nd International Conference on Machine Learning},
  pages = 	 {2256--2265},
  year = 	 {2015},
  volume = 	 {37},
  series = 	 {Proceedings of Machine Learning Research},
  address = 	 {Lille, France},
  month = 	 {07--09 Jul},
  publisher =    {PMLR},
  url = 	 {https://proceedings.mlr.press/v37/sohl-dickstein15.html}
}

@inproceedings{chen2023diffusiondet,
  title={Diffusiondet: Diffusion model for object detection},
  author={Chen, Shoufa and Sun, Peize and Song, Yibing and Luo, Ping},
  booktitle={Proceedings of the IEEE/CVF International Conference on Computer Vision},
  pages={19830--19843},
  year={2023}
}

@inproceedings{kong2020diffwave,
  title={DiffWave: A Versatile Diffusion Model for Audio Synthesis},
  author={Kong, Zhifeng and Ping, Wei and Huang, Jiaji and Zhao, Kexin and Catanzaro, Bryan},
  booktitle={International Conference on Learning Representations},
  year={2020}
}

@inproceedings{li2023simple,
  title={Simple Approach for Aspect Sentiment Triplet Extraction Using Span-Based Segment Tagging and Dual Extractors},
  author={Li, Dongxu and Yang, Zhihao and Lan, Yuquan and Zhang, Yunqi and Zhao, Hui and Zhao, Gang},
  booktitle={Proceedings of the 46th International ACM SIGIR Conference on Research and Development in Information Retrieval},
  pages={2374--2378},
  year={2023}
}

@inproceedings{gao2021question,
  title={Question-driven span labeling model for aspect--opinion pair extraction},
  author={Gao, Lei and Wang, Yulong and Liu, Tongcun and Wang, Jingyu and Zhang, Lei and Liao, Jianxin},
  booktitle={Proceedings of the AAAI conference on artificial intelligence},
  volume={35},
  number={14},
  pages={12875--12883},
  year={2021}
}

@article{cao2021image,
  title={The image local autoregressive transformer},
  author={Cao, Chenjie and Hong, Yuxin and Li, Xiang and Wang, Chengrong and Xu, Chengming and Fu, Yanwei and Xue, Xiangyang},
  journal={Advances in Neural Information Processing Systems},
  volume={34},
  pages={18433--18445},
  year={2021}
}

@inproceedings{he2023diffusionbert,
  title={DiffusionBERT: Improving Generative Masked Language Models with Diffusion Models},
  author={He, Zhengfu and Sun, Tianxiang and Tang, Qiong and Wang, Kuanning and Huang, Xuan-Jing and Qiu, Xipeng},
  booktitle={Proceedings of the 61st Annual Meeting of the Association for Computational Linguistics},
  pages={4521--4534},
  year={2023}
}

@article{vaswani2017attention,
  title={Attention is all you need},
  author={Vaswani, Ashish and Shazeer, Noam and Parmar, Niki and Uszkoreit, Jakob and Jones, Llion and Gomez, Aidan N and Kaiser, {\L}ukasz and Polosukhin, Illia},
  journal={Advances in neural information processing systems},
  volume={30},
  year={2017}
}

@article{han2023information,
  title={Is information extraction solved by chatgpt? an analysis of performance, evaluation criteria, robustness and errors},
  author={Han, Ridong and Peng, Tao and Yang, Chaohao and Wang, Benyou and Liu, Lu and Wan, Xiang},
  journal={arXiv preprint arXiv:2305.14450},
  year={2023}
}

@inproceedings{devlin-etal-2019-bert,
    title = "{BERT}: Pre-training of Deep Bidirectional Transformers for Language Understanding",
    author = "Devlin, Jacob  and
      Chang, Ming-Wei  and
      Lee, Kenton  and
      Toutanova, Kristina",
    booktitle = "Proceedings of the 2019 Conference of the North {A}merican Chapter of the Association for Computational Linguistics",
    year = "2019",
    publisher = "Association for Computational Linguistics",
    pages = "4171--4186",
}

@inproceedings{kou2023bayesdiff,
  title={BayesDiff: Estimating Pixel-wise Uncertainty in Diffusion via Bayesian Inference},
  author={Kou, Siqi and Gan, Lei and Wang, Dequan and Li, Chongxuan and Deng, Zhijie},
  booktitle={The Twelfth International Conference on Learning Representations},
  year={2023}
}

@article{du2024diffusion,
  title={Diffusion-Based Probabilistic Uncertainty Estimation for Active Domain Adaptation},
  author={Du, Zhekai and Li, Jingjing},
  journal={Advances in Neural Information Processing Systems},
  volume={36},
  year={2024}
}

@article{sun2025senticnet,
  title={SenticNet and Abstract Meaning Representation driven Attention-Gate semantic framework for aspect sentiment triplet extraction},
  author={Sun, Xiaowen and Qi, Jiangtao and Zhu, Zhenfang and Li, Meng and Meng, Jing},
  journal={Engineering Applications of Artificial Intelligence},
  volume={139},
  pages={109625},
  year={2025},
  publisher={Elsevier}
}

@inproceedings{peng2024prompt,
  title={Prompt based tri-channel graph convolution neural network for aspect sentiment triplet extraction},
  author={Peng, Kun and Jiang, Lei and Peng, Hao and Liu, Rui and Yu, Zhengtao and Ren, Jiaqian and Hao, Zhifeng and Yu, Philip S},
  booktitle={Proceedings of the 2024 SIAM International Conference on data mining (SDM)},
  pages={145--153},
  year={2024},
  organization={SIAM}
}

@article{li2024improving,
  title={Improving span-based aspect sentiment triplet extraction with part-of-speech filtering and contrastive learning},
  author={Li, Qingling and Wen, Wushao and Qin, Jinghui},
  journal={Neural Networks},
  volume={177},
  pages={106381},
  year={2024},
  publisher={Elsevier}
}

@article{zhang2025aspect,
  title={Aspect Sentiment Triplet Extraction with Syntax-Semantics Graph Convolutional Network},
  author={Zhang, Jingyun and Xu, Shuwei and Gao, Xin and Tang, Zhiwei},
  journal={International Journal of Computational Intelligence Systems},
  volume={18},
  number={1},
  pages={167},
  year={2025},
  publisher={Springer}
}

@article{lu2025qaie,
  title={QAIE: LLM-based quantity augmentation and information enhancement for few-shot aspect-based sentiment analysis},
  author={Lu, Heng-yang and Liu, Tian-ci and Cong, Rui and Yang, Jun and Gan, Qiang and Fang, Wei and Wu, Xiao-jun},
  journal={Information Processing\&Management},
  volume={62},
  number={1},
  pages={103917},
  year={2025},
  publisher={Elsevier}
}

@article{yi2025diffusyn,
  title={DiffuSyn: A Diffusion-Driven Framework With Syntactic Dependency for Aspect Sentiment Triplet Extraction},
  author={Yi, Qiuhua and Kong, Xiangjie and Zhu, Linan and Zhang, Chenwei and Shen, Guojiang},
  journal={IEEE Transactions on Audio, Speech and Language Processing},
  year={2025},
  publisher={IEEE}
}

@inproceedings{liu2024let,
  title={Let’s rectify step by step: Improving aspect-based sentiment analysis with diffusion models},
  author={Liu, Shunyu and Zhou, Jie and Zhu, Qunxi and Chen, Qin and Bai, Qingchun and Xiao, Jun and He, Liang},
  booktitle={Proceedings of the 2024 joint international conference on computational linguistics, language resources and evaluation (LREC-COLING 2024)},
  pages={10324--10335},
  year={2024}
}

@article{lin2025revisiting,
  title={Revisiting aspect sentiment triplet extraction: A span-level approach with enhanced contextual interaction},
  author={Lin, Run and Gan, Yanglei and Lan, Tian and Liu, Xueyi and Cai, Yuxiang and Li, Changlin and Luo, Da and Liu, Qiao},
  journal={Expert Systems with Applications},
  volume={268},
  pages={126126},
  year={2025},
  publisher={Elsevier}
}

@article{gan2025optimizing,
  title={Optimizing boundary dynamics for nested named entity recognition via semantic refinement and trimming},
  author={Gan, Yanglei and Liu, Yao and Cai, Yuxiang and Lin, Run and Yang, Song and Liu, Qiao and Wang, Yashen and Shi, Xiaojun},
  journal={Neural Networks},
  pages={108218},
  year={2025},
  publisher={Elsevier}
}

@inproceedings{gan2026negative,
  title={Negative-Aware Diffusion Process for Temporal Knowledge Graph Extrapolation},
  author={Gan, Yanglei and He, Peng and Cai, Yuxiang and Lin, Run and Zhou, Guanyu and Liu, Qiao},
  booktitle={Findings of the Association for Computational Linguistics: EACL 2026},
  pages={3352--3367},
  year={2026}
}

@inproceedings{cai2024predicting,
  title={Predicting the unpredictable: uncertainty-aware reasoning over temporal knowledge graphs via diffusion process},
  author={Cai, Yuxiang and Liu, Qiao and Gan, Yanglei and Li, Changlin and Liu, Xueyi and Lin, Run and Luo, Da and JiayeYang, JiayeYang},
  booktitle={Findings of the Association for Computational Linguistics: ACL 2024},
  pages={5766--5778},
  year={2024}
}
%




\end{document}